\documentclass{article}

 \PassOptionsToPackage{numbers, compress}{natbib}
\usepackage[preprint]{nips_2018}

\usepackage[utf8]{inputenc} % allow utf-8 input
\usepackage[T1]{fontenc}    % use 8-bit T1 fonts
\usepackage{hyperref}       % hyperlinks+++
\usepackage{url}            % simple URL typesetting
\usepackage{booktabs}       % professional-quality tables
\usepackage{amsfonts}       % blackboard math symbols
\usepackage{nicefrac}       % compact symbols for 1/2, etc.
\usepackage{microtype}      % microtypography
\usepackage{amsmath}
\usepackage{amscd}
\usepackage{amsfonts}
\usepackage{amssymb}
\usepackage{amsthm}
\usepackage[dvips]{color}
\usepackage{epsfig}
\usepackage{graphics}
\usepackage[all]{xy}
\usepackage{multicol}
\usepackage{multirow}
\usepackage{tabularx}
 \theoremstyle{definition}
 
 \theoremstyle{remark}

\title{Matrix embedding method in match for session-based recommendation}

\author{
  Qizhi Zhang  \\
  \texttt{qizhi.zqz@alibaba-inc.com} \\
 \And
 Yi Lin \\
 \texttt{bani.ly@alibaba-inc.com}\\
 \And
 Kangle Wu \\
 \texttt{ningci.wkl@antfin.com} \\
 \And
 Yongliang Li \\
  \texttt{anthonylee.liyl@tmall.com} \\
  \And
  Anxiang Zeng \\
 \texttt{renzhong@taobao.com} \\
}

\begin{document}
 %\nipsfinalcopy is no longer used

\maketitle

\begin{abstract}
Session based model is widely used in recommend system. It use the user click sequence as input of a
 Recurrent Neural Network (RNN), and get the output of the RNN network as the vector embedding of the session, and use the inner product
 of the vector embedding of session and the vector embedding of the next item as the score that is the metric of the interest to the next item.
 This method can be used for the "match" stage for the recommendation system whose item number is very big by using some index method like KD-Tree or Ball-Tree and etc.. But this method repudiate the variousness
 of the interest of user in a session. We generated the model to modify the vector embedding of session to a symmetric matrix embedding, that is equivalent to a quadratic form on the 
 vector space of items. The score is builded as the value of the vector embedding of next item under the quadratic form. The eigenvectors of the symmetric matrix embedding corresponding to the positive eigenvalues are conjectured to represent the interests of user in the session. This method can be used for the "match" stage also. The experiments show that this method is better than the method of vector embedding.

\end{abstract}

\section{Introduction}

In some large E-COMMERCE, for example TABAO, AliExpress, the recommend algorithm is  divided into two stages, i.e. the stage "match" and the stage "rank".
In the stage "match", we need  select item set with size $O(10^2 \sim 10^3)$ from all the items. In the stage "rank", we compute a score for the items in the match item set, and rank them by
score.  We can use model of any form in the stage rank, but there are some  restriction for the model in the stage match,
%The main requirement of algorithm in "match" 
it is that it must can  \textbf{quick} pick up  $O(10^2 \sim 10^3)$ items from $O(10^8)$ or more items, hence it need an index. 
Only the models can generate index can be used in the stage match. The most familiar
model for match is the static model, it compute the conditional probability $p_{i,j}=P(\mbox{view } i | \mbox{view } j)$ as score, and save a table with the fields "triger id", "item id", "score" indexed by "triger id" and "score" in offline. In online, we recall items with top N score using "triger id" as index, where "triger id" com from the items which user has behaviour.

Sequence prediction is a problem that involves using historical sequence information to predict the next value or values in the sequence.
There are a lot of  applications of this type, for example,  the language model and recommend 
 system.
 Recurrent Neural Network (RNN) is widely used to solve the sequence prediction  
 problems.

 For a given sequence $x_1, x_2, \cdots, x_n$, we wish predict $x_{n+1}$. In the 
 situation of recommend system, the $x_i$s are the item which user clicked ( or buy, added to wish list, 
 etc), hence the sequence $x_1, x_2, \cdots, x_n$ is the representative of user. 
 In the situation of language model, the $x_i$s are the words in the sentence, hence the sequence $x_1, x_2, \cdots, x_n$ is the representative of front part
 of the sentence. 
 
 The final layer is a full connectional layer with a softmax.

\begin{figure}[p]
  \begin{minipage}[t]{0.3\linewidth}
 \includegraphics[scale=0.3]{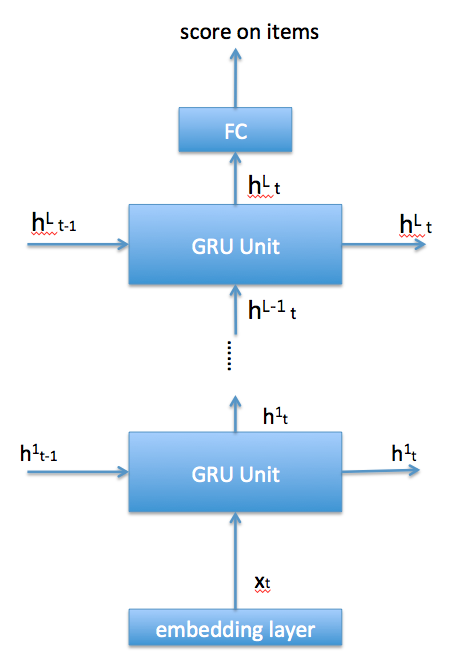}
\caption{The network structure in \cite{SESSION_BASED}}
\label{FC}
\end{minipage}
\hfill
\begin{minipage}[t]{0.6\linewidth}
  \includegraphics[scale=0.45]{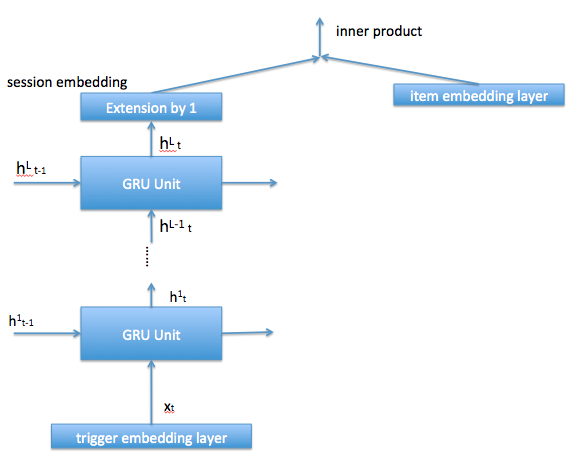}
\caption{A equivalent network structure to \cite{SESSION_BASED}}
\label{vec_embed}
\end{minipage}
\end{figure}

In \cite{SESSION_BASED}, a session based model for recommend system is proposed. 
The model is like Figure \ref{FC}.

This network structure is equivalent to the network structure in Figure 
\ref{vec_embed}. In Figure \ref{vec_embed}, we give two embedding for every item: if
the item is passed, we call it ``trigger'', and call the embedding ``trigger embedding''; if the item is to be predict score,
we call it ``item'', and call its embedding ``item embedding''. 
The layer ``Extension by 1'' means the static map
\[
\begin{array}{rcl}
  \mathbb{R}^{(n)} & \longrightarrow & \mathbb{R}^{(n+1)}  \\
  (a_1, \cdots, a_n)^T & \mapsto & (a_1, \cdots, a_n, 1)^T
\end{array}
\]

Because the output $h^L_t$ of final GRU layer collected all the trigger information up to time $t$ of the session, we can 
view the output of layer ``Extension by 1'' as ``session embedding''.  We set the dimension of item embedding equal to the dimension of session embedding, 
and define the output of network as the inner product of the session embedding 
and the item embedding.
%\[
%\begin{array}{ccccl}
%   \mathbb{R}^n & \times & \mathbb{R}^{n+1} & \longrightarrow 
%  & \mathbb{R} \\
%  ((a_1, \cdots a_n)& , & (b_1, \cdots, b_{n+1})) & \mapsto & \sum_{i=1}^n a_ib_i +b_{n+1}
%\end{array}
%\] 
It easy to see that the network structure in Figure\ref{FC} and Figure \ref{vec_embed} 
are equivalent under the corresponding 

\[
\begin{array}{rcl}
    \mbox{FC layer} & \longrightarrow &  \mbox{item embedding layer}   \\
     (x \mapsto \mathop{softmax}(Ax+b))   &  \mapsto &   (i \mapsto  \mathop{concat}(A_i, b_i) )
\end{array}
\] 
where $A_i$ means the $i$ row of the matrix A and $b_i$ means the $i$ element of the column vector $b$.  
Hence we call this method ``vector embedding method''.

The session based model of with vector embedding method can used as a model for match. In fact, after the model is trained, 
we can  save the vector embedding of items with some index, for example, KD-Tree, BallTree, .... When a user visit our recommend page, we compute the vector embedding $x$
of users session using the click sequences of user, and find the Top N items which vector embedding has max inner product with $x$ using index.

\begin{figure}[p]
  \begin{minipage}[t]{0.3\linewidth}
 \includegraphics[scale=0.4]{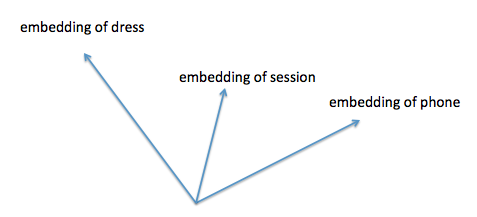}
\caption{The inherent defect of vector embedding method}
\label{inherent_defect}
\end{minipage}
\hfill
\begin{minipage}[t]{0.5\linewidth}
  \includegraphics[scale=0.36]{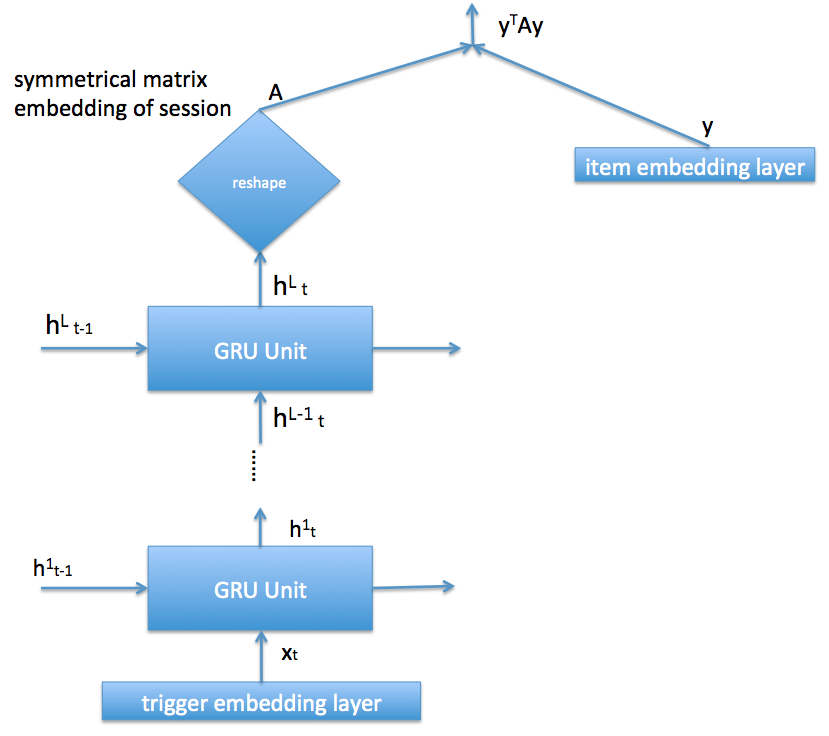}
\caption{The matrix embedding method}
\label{mat_embed}
\end{minipage}
\end{figure}

But the vector embedding method has an inherent defect.
Because the interests of user may not be single.
 Suppose
there are the item embeddings of dress and phone as shown in (Figure 
\ref{inherent_defect}).
Generally the interest to dress is independent to the interest to phone, we can 
suppose
they are linear independent.
 If a user clicked  dresses 20 times and phones 10 times
in one session, then the vector embedding of this session will 

mainly try to close to 
the dress, but will be drag away by phone under training, in the result, the vector embedding of session will lie 
between the dress and phone, which is not close to neither dress or phone. 
Hence when we predict using this embedding, we will recommend something like comb of dress and phone to the user
as the top 1 selection, 
instead of the most interested dress. 
In other words, the scheme of vector embedding deprived the variousness of the intersection user in one session.

In order to model the variousness of the intersection user in one session, we 
use ``matrix embedding'' of session instead of ``vector embedding''.

In our method, the items are modeled as vectors of dimension $n$ still, but a session is modeled
as a symmetric matrix in $M_{n}(\mathbb{R})$ instead of a vector in $\mathbb{R}^{(n)}$.  The score
which represent the interest of the session to the item is modeled as 
\[
y^TAy,
\]
where $y$ is the vector embedding of item, and $A$ is the matrix embedding of the session.
Because the symmetric matrix $A$ has the eigendecomposition (\cite{Eigendecomposition})
\[
A=Q \Lambda Q^T,
\]
where $Q$ is a real orthogonal square matrix, and $\Lambda$ is a diagonal square matrix 
with the elements $\lambda_1 \geq \lambda _2 \cdots \geq \lambda _n$ on the diagonal 
line. In fact, $\lambda_1 , \lambda _2 \cdots , \lambda _n$ are the eigenvalues 
of $A$, and the $i$-th column of $Q$ is the eigenvector according to $\lambda _i$. 
In the example in Figure \ref{inherent_defect}, the matrix embedding of session 
can has two  eigenvalues $\lambda_1 > \lambda _2$  significantly greater than others, whose eigenvectors are 
close
to the lines along the embedding vector of dress and the embedding vector of phone respectively.
Hence, the function
\[
\begin{array}{rcl}
  U_1 ^n(0) & \longrightarrow & \mathbb{R} \\
  y & \mapsto & y^TAy
\end{array}
\]
take its max value close to the direction of dress, where $U_r(0)$ means the 
unit ball in $\mathbb{R}^{(n)}$. When we using this model to predict, we will 
recommend dress to the user as the top 1 selection.

\section{Network structure}

The Network 
structure of our new method is showed in Figure \ref{mat_embed}. The main 
difference between Figure \ref{mat_embed} and Figure \ref{vec_embed} is that 
\begin{enumerate}
\item We set the dimension of the hidden layers to be $\frac{n(n+1)}{2}$, where $n$ is the dimension of 
the embedding vectors of items. 
\item We use the layer ``reshape to a symmetric matrix'' instead of the  layer ``extension by 
1''. The layer ``reshape to a symmetric matrix'' is defined as

\[
\begin{array}{rcl}
  \mathbb{R}^{\frac{n(n+1)}{2}} & \longrightarrow & M_n(\mathbb{R}) \\
  (z_i)_{i=1} ^{\frac{n(n+1)}{2}} & \mapsto &  (a_{i,j}) _{i,j =1} ^n 
\end{array},
\]
where $a_{i,j}=
\left\{
\begin{array}{ll}
  z_{\frac{i(i-1)}{2}+j} & \mbox{ if } i \leq j \\
  a_{j,i} & \mbox{ otherwise}.
\end{array}
\right.$

\item We use the layer
\[
\begin{array}{ccccl}
  M_n(\mathbb{R}) & \times & \mathbb{R}^{(n)} & \longrightarrow & \mathbb{R} \\
  (A &,& y) & \mapsto & y^TAy
\end{array}
\]
as the score layer instead of the inner product.

\item There is a modifying of the item embedding layer. It is the upper half hyperplane 
embedding, i.e, the embedding vectors of items in the  upper half hyperplane 
\[
\mathbb{H}^{(n)} := \{ (y_1, \cdots y_n)^T \in \mathbb{R}^{(n)} : y_n>0 \}.
\]
This modifying improve the performance greatly. We give some illustration of the reason: 
because the score value $y^TAy$ is invariant under the  transformation 
$y \mapsto -y$, and if we train the model without the modifying of the item 
embedding layer, the embedding of items will lost its direction in training. 
The realizing of the layer ``upper half hyperplane 
embedding'' can be got through apply $\exp$ to the final coordinate of a vector in ordinal embedding layer.  
\end{enumerate}

\section{Index method}

For using as match method, we  give two index method of matrix embedding method. 

We formulate the problem as following:

There a lot of vector $\{x_i\}_i \subset \mathbb{R}^{(n)}$, for a  symmetrical matrix $A \in M_n(\mathbb{R})$, how we can find the top N $x_i$s such that
$x^TAx$ is maximal.

\subsection{Flatten}
Because $A$ is a symmetrical matrix, we have
\[
x^TAx=\sum _{i \leq j} a_{i,j}  k_{i,j}x_ix_j =<\Gamma _1(A), \Gamma _2(x)>
\]
where $k_{i,j}= \left\{ 
\begin{array} {ll}
1 & \mbox{ if } i=j \\
2 & \mbox{ otherwise}
\end{array}
\right.
$, and $\Gamma _1 (A):=( a_{i,j} ) _{i \leq j}$,  $\Gamma _2 (x):=(k_{i,j} x_ix_j) _{i \leq j} $. Therefore, we can map the user session matrix embedding $A$ into a linear space of dimension $R^{\frac{n(n+1)}{2}}$ using $\Gamma_1$, and map the item vector embedding $x$ into the same linear space using $\Gamma _2$, the score $x^TAx$ is equal to the inner product
of $\Gamma _1(A)$ and $\Gamma _2(x)$. Hence, we can construct the index of $\Gamma _2 (x)$ for the vector embedding $x$ for all items in offline and get Top N items of maximal inner product for every 
$\Gamma _1 (A)$ in online
 like usual method to get Top N items of maximal  inner product.

\subsection{Decomposition}  
The Flatten method need build index for vectors of dimension $\frac{n(n+1)}{2}$. When the dimension $n$ is big, it is difficult to save the data, build the index and search the items of maximal inner product. Hence we need a  method to get the approximate top N items of maximal  inner product faster.  In fact, we have the Singular Value Decomposition 
\[
A=\sum _{i=1} ^ n \lambda _i \alpha _i \alpha _i ^T
\]
where $\lambda _1 \geq \lambda _2  \geq \cdots \geq \lambda _n$, and $\alpha _i \in \mathbb{H}^{(n)}$.
Hence we have 
\[
x^TAx=\sum _{i=1} ^n \lambda _i <\alpha _i, x>^2
\]
As a approximate method, we take a small positive integral number $k$,  and take the top N items of maximal  inner product $<\alpha _i, x>$ for $i=1,2, \cdots, k$, and hence we have $kN$ items, then we take top N items from these kN items by computing $x^T Ax$.

\section{Experiments}

We give the experiment to compare matrix embedding method and vector embedding method on the Dataset RSC15 (RecSys Challenge 2015 \footnote{ 
  http://2015.recsyschallenge.com/}) and the last.fm \cite{last_fm} dataset .

%To compare with the original GRU4REC model, here we performed the symmetric matrix model on the ACM RecSys 2015 Challenge dataset used in the original paper of GRU4REC [1] and followed the same preprocessing and split procedure described in [1].

%Furthermore, in order to prove that our improvement of GRU4REC can preform well in different scenarios, we also train and evaluate the symmetric matrix model on another music playlists dataset  last.fm \cite{last_fm}.

%The main work of Hidasi et al. in [1] is training the GRU4REC model on the previous session-based click streams, and try to predict the next click item of each session in each time step. They evaluate the performance of the model by checking the rank of the predict score for the target item and calculate the average hit rate (recall@20) and the average Mean Reciprocal Rank (mrr@20) .

For the  RSC15 dataset, after tuning  the hyperparameters on the validation set, we retrained the three models above on the whole days among six months, and used the last single day to evaluate those models. When it comes to the last.fm playlists dataset, since the playlists have no timestamps,  we followed the preprocessing procedure of \cite{last_fm_paper}, that is, randomly assigned each playlist to one of the 31 buckets (days), and used the lastest single day to evaluate.

We compare the tree models:

\textbf{GRU4REC} We re-implemented the code of GRU4REC which Hidasi et al. released online \cite{SESSION_BASED} in Tensorflow framework, including the whole GRU4REC architecture, the training procedure as well as the evaluation procedure.

\textbf{GRU4REC with symmetric matrix} To address the problem of GRU4REC demonstrated in section 1, we replace the output of the GRU i.e. the embedding vector of the current session with a symmetric matrix. More specifically...

\textbf{GRU4REC with fully connected layer} In addition to the above models, we also create a controlled experiment model as shown in Figure 3, which mainly based on the GRU4REC model but only add a fully connected layer right after the output of the GRU to expands the embedding vector size of the GRU output from $n$ dimensions to $n(n+1)/2$ dimensions.

\subsection{ACM RecSys 2015 Challenge Dataset}

In order to evaluate the performance of the three models described in section 2.1, we constrained the total quantity of their parameters to the same range. The detail of the networks' architecture are shown in table 1 respectively.

Table 2 shows the results when testing those three models on the last day of the ACM RecSys 2015 Challenge dataset for 10 epochs. After tuning on the validation set, we set lr=0.002, batch size = 256 for all the experiments. And since the GRU4REC and GRU4REC with FC layer model have less hidden units, dropout=0.8 shows better performance for them while dropout=0.5 performs better for the symmetric matrix model. Meanwhile, they're using bpr loss and adam optimizer in all cases.

\begin{table}
\centering
\caption{Results for the RSC15 dataset.}
\begin{tabularx}{13cm}{XXX}  
\hline                      
Method & recall@20  &  mrr@20  \\
\hline
GRU4REC  & 0.389 & 0.135 \\
GRU4REC+FC  & 0.515 & 0.515 \\
GRU4REC+Matrix  & {\bf 0.749} & {\bf 0.748} \\
GRU4REC(1000)  & 0.632 & 0.247 \\
\hline
\end{tabularx}
\end{table}

We additionally include the results in \cite{SESSION_BASED} which uses 1000 hidden units for the GRU4REC model. 
It's obvious that by combining symmetric matrix embedding method with GRU4REC, we could use less parameter to achive better recall@20 and mrr@20 performance.

\renewcommand{\arraystretch}{1.5} 
\begin{table}[tp]
  \centering
  \fontsize{6.5}{8}\selectfont
  \caption{Network Parameters For RecSys15 Dataset.}
  \label{tab:RecSys15_params}
    \begin{tabular}{|c|c|c|c|c|c|c|c|c|c|}
    \hline
    \multirow{2}{*}{Model}&
    \multicolumn{3}{c|}{GRU4REC}&\multicolumn{3}{c|}{GRU4REC+FC}&\multicolumn{3}{c|}{GRU4REC+Matrix}\cr\cline{2-10}
    &shape&params&total&shape&params&total&shape&params&total\cr
    \hline
    \hline
    {input\_embedding}&{(37958, 32)}&1214656&1214656&{(37958, 32)}&1214656&1214656&{(37958, 32)}&1214656&1214656\cr\hline
    {softmax\_W}&{(37958, 64)}&2429312&3643968&{(37958, 55)}&2087690&3302346&{(37958, 32)}&1214656&2429312\cr\hline
    {softmax\_b}&{(37958,)}&37958&3681926&{(37958,)}&37958&3340304&-&-&-\cr\hline
    {gru\_cell/dense/kernel}&-&-&-&{(10, 55)}&550&3340854&-&-&-\cr\hline
    {gru\_cell/dense/bias}&-&-&-&{(55,)}&55&3340909&-&-&-\cr
    \hline
    {gru\_cell/gates/kernel}&{(96, 128)}&12288&3694214&{(42, 20)}&840&3341749&{(560, 1056)}&591360&3020672\cr\hline
    {gru\_cell/gates/bias}&(128,)&128&3694342&{(20,)}&20&3341769&{(1056,)}&1056&3021728\cr\hline
    {gru\_cell/candidate/kernel}&(96, 64)&6144&3700486&{(42, 10)}&420&3342189&{(560, 528)}&295680&3317408\cr\hline
    {gru\_cell/candidate/bias}&(64,)&64&{\bf 3700550}&{(10,)}&10&{\bf 3342199}&{(528,)}&528&{\bf 3317936}\cr
    \hline
    \end{tabular}
\end{table}

\subsection{Last.FM playlists Dataset}
For the last.fm music playlists datasets, we applied the same network structure for each model as mentioned above, the specific parameters are shown in Table 3 as follows.

\renewcommand{\arraystretch}{1.5} 
\begin{table}[tp]
  \fontsize{6.5}{8}\selectfont
  \caption{Network Parameters For Last.fm Dataset.}
  \centering
  \label{tab:playlists_params}
    \begin{tabular}{|c|c|c|c|c|c|c|c|c|c|}
    \hline
    \multirow{2}{*}{Model}&
    \multicolumn{3}{c|}{GRU4REC}&\multicolumn{3}{c|}{GRU4REC+FC}&\multicolumn{3}{c|}{GRU4REC+Matrix}\cr\cline{2-10}
    &shape&params&total&shape&params&total&shape&params&total\cr
    \hline
    \hline
    {input\_embedding}&{(200668, 32)}&6421376&6421376&{(200668, 32)}&6421376&6421376&{(200668, 32)}&6421376&6421376\cr\hline
    {softmax\_W}&{(200668, 64)}&12842752&19264128&{((200668, 55)}&11036740&17458116&{(200668, 32)}&6421376&12842752\cr\hline
    {softmax\_b}&{((200668,)}&200668&19464796&{(200668,)}&200668&17658784&-&-&-\cr\hline
    {gru\_cell/dense/kernel}&-&-&-&{(10, 55)}&550&17659334&-&-&-\cr\hline
    {gru\_cell/dense/bias}&-&-&-&{(55,)}&55&17659389&-&-&-\cr
    \hline
    {gru\_cell/gates/kernel}&{(96, 128)}&12288&19477084&{(42, 20)}&840&17660229&{(560, 1056)}&591360&13434112\cr\hline
    {gru\_cell/gates/bias}&(128,)&128&19477212&{(20,)}&20&17660249&{(1056,)}&1056&13435168\cr\hline
    {gru\_cell/candidate/kernel}&(96, 64)&6144&19483356&{(42, 10)}&420&17660669&{(560, 528)}&295680&13730848\cr\hline
    {gru\_cell/candidate/bias}&(64,)&64&{\bf 19483420}&{(10,)}&10&{\bf 17660679}&{(528,)}&528&{\bf 13731376}\cr
    \hline
    \end{tabular}
\end{table}

Since the music playlists dataset is so different from the e-commerce click sequence dataset, after tuning on the validation set, we finally set lr = 0.0012 for all cases while the batch size and dropout configuration remain the same.

Table 4 shows results for the last.fm music playlists datasets. We can notice the same trend when comparing the results with the RecSys15 dataset.

\renewcommand\arraystretch{1.5}  % why doesn't work??
\begin{table}
\centering
\caption{Results for the last.fm dataset.}
\begin{tabularx}{13cm}{XXX}  
\hline                      
Method & recall@20  &  mrr@20  \\
\hline
GRU4REC  & 0.027 & 0.022 \\
GRU4REC+FC  & 0.054 & 0.054 \\
GRU4REC+Matrix  & {\bf 0.164} & {\bf 0.164} \\
GRU4REC(1000)  & 0.121 & 0.053 \\
\hline
\end{tabularx}
\end{table}


\begin{thebibliography}{99}
 \bibitem{SESSION_BASED} Balazs Hidasi, Alexandros Karatzoglou, Linas Baltrunas, and Domonkos Tikk. 2016. 
 Session-based Recommendations with Recurrent Neural Networks. International Conference on Learning Representations(2016). %http://arxiv.org/abs/1511. 06939
 
 \bibitem{top_k} Balazs Hidasi, Alexandros Karatzoglou. Recurrent Neural Networks with Top-k Gains for Session-based 
 Recommendations.
 CIKM '18 Proceedings of the 27th ACM International Conference on Information and Knowledge Management
Pages 843-852. Torino, Italy — October 22 - 26, 2018 
ACM New York, NY, USA ©2018 
table of contents ISBN: 978-1-4503-6014-2. 
 
 \bibitem{DWell} Bogina, Veronika, and Tsvi Kuflik. “Incorporating dwell time in session-based recommendations with recurrent Neural networks.” CEUR Workshop Proceedings. Vol. 1922. 2017.

\bibitem{Personalizing}
Quadrana, Massimo, et al. “Personalizing session-based recommendations with hierarchical recurrent neural networks.” Proceedings of the Eleventh ACM Conference on Recommender Systems. ACM, 2017.
 
 
 \bibitem{Eigendecomposition} https://en.wikipedia.org/wiki/Eigendecomposition\_of\_a\_matrix
 
 \bibitem{last_fm}Thierry Bertin-Mahieux and Daniel P.W. Ellis and Brian Whitman and Paul Lamere. The Million Song Dataset. Proceedings of the 12th International Conference on Music Information Retrieval ({ISMIR} 2011). http://millionsongdataset.com/lastfm/
 
 \bibitem{last_fm_paper}Dietmar Jannach and Malte Ludewig. When Recurrent Neural Networks meet the Neighborhood for Session-Based Recommendation. RecSys '17 Proceedings of the Eleventh ACM Conference on Recommender Systems
Pages 306-310.
 
 %@INPROCEEDINGS{Bertin-Mahieux2011,
  %author = {Thierry Bertin-Mahieux and Daniel P.W. Ellis and Brian Whitman and Paul Lamere},
  %title = {The Million Song Dataset},
  %booktitle = {{Proceedings of the 12th International Conference on Music Information
%	Retrieval ({ISMIR} 2011)}},
  %year = {2011},
  %owner = {thierry},
  %timestamp = {2010.03.07}
%}
 
 
 \end{thebibliography}
\end{document}